\pdfoutput=1
\documentclass[11pt]{article}

\usepackage[]{ACL2023}

\usepackage{times}
\usepackage{latexsym}

\usepackage[T1]{fontenc}
\usepackage[utf8]{inputenc}

\usepackage{microtype}

\usepackage{inconsolata}

\usepackage{graphicx}
\usepackage{bm}
\usepackage{amsmath}
\usepackage{amsfonts}
\usepackage{bbm}

\usepackage{helvet}
\usepackage{courier}
\usepackage{multirow}
\usepackage{bm}
\usepackage{amsmath}
\usepackage{mdwlist}

\usepackage{relsize}
\usepackage{amsfonts}
\usepackage{url}
\usepackage{graphicx}
\usepackage{subfigure}
\usepackage{caption}
\usepackage{hhline}
\usepackage{color}
\usepackage{colortbl}
\usepackage{booktabs}
\usepackage{tabularx}
\usepackage{amsmath}
\usepackage{makecell}
\usepackage[linesnumbered,boxed,ruled,commentsnumbered]{algorithm2e}
\usepackage{lipsum}
\usepackage{booktabs}
\usepackage{pifont}
\title{Open Set Relation Extraction via Unknown-Aware Training}

\author{Jun Zhao$^{1}$\footnotemark[1],\ \ Xin Zhao$^{1}$\footnotemark[1],\ \ Wenyu Zhan$^{1}$,\ \ Qi Zhang$^{1}$,\ \ Tao Gui$^{2}$\footnotemark[2], \\ \textbf{Zhongyu Wei}$^3$\textbf{,}\ \ \textbf{Yunwen Chen}$^4$\textbf{,}\ \ \textbf{Xiang Gao}$^4$\textbf{,}\ \ \textbf{Xuanjing Huang}$^{1,5}$\footnotemark[2]\\
  $^1$School of Computer Science, Fudan University\\
  $^2$Institute of Modern Languages and Linguistics, Fudan University\\
  $^3$School of Data Science, Fudan University\\
  $^4$DataGrand Information Technology (Shanghai) Co., Ltd.\\
  $^5$International Human Phenome Institutes (Shanghai)\\
  \texttt{\{zhaoj19,qz,tgui\}@fudan.edu.cn, zhaoxin21@m.fudan.edu.cn}}

\begin{document}
\maketitle
\renewcommand{\thefootnote}{\fnsymbol{footnote}}
\footnotetext[1]{Equal Contributions.}
\footnotetext[2]{Corresponding authors.}
\begin{abstract}
The existing supervised relation extraction methods have achieved impressive performance in a closed-set setting, where the relations during both training and testing remain the same. In a more realistic open-set setting, unknown relations may appear in the test set. Due to the lack of supervision signals from unknown relations, a well-performing closed-set relation extractor can still confidently misclassify them into known relations. In this paper, we propose an unknown-aware training method, regularizing the model by dynamically synthesizing negative instances. 
To facilitate a compact decision boundary, 
``difficult'' negative instances are necessary. Inspired by text adversarial attacks, we adaptively apply small but critical perturbations to original training instances and thus synthesizing negative instances that are more likely to be mistaken by the model as known relations.
Experimental results show that this method achieves SOTA unknown relation detection without compromising the classification of known relations.
\end{abstract}

\section{Introduction}
    \begin{figure}[t]
        \includegraphics[width=\columnwidth]{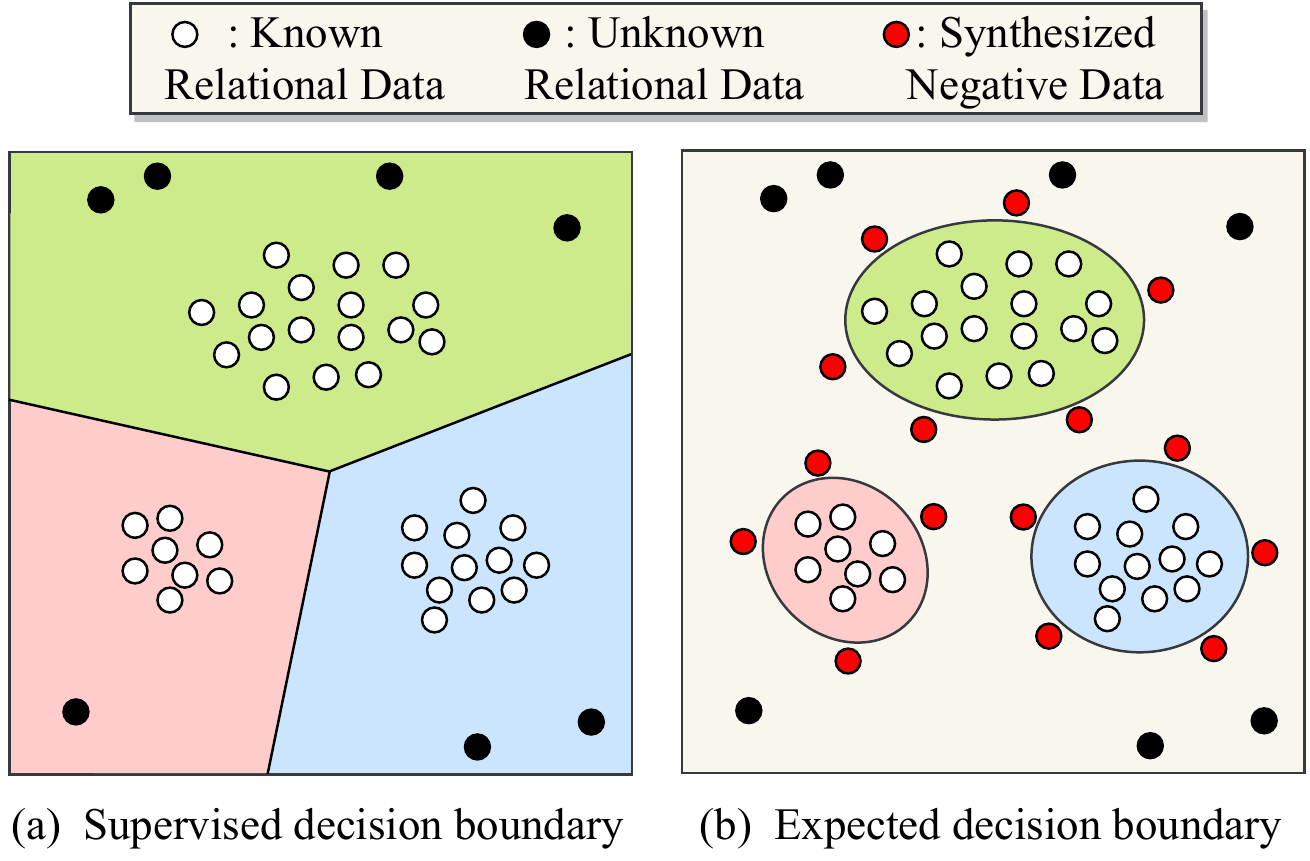}
        \caption{The decision boundary optimized only on the known relations
        cannot cope with an open set setting, in which the input may come from the relations unobserved in training. We target at regularizing the decision boundary by synthesizing difficult negative instances.}
        \label{fig:intro}
    \end{figure}
Relation extraction (RE) is an important basic task in the field of natural language processing, aiming to extract the relation between entity pairs from unstructured text. The extracted relation facts have a great practical interest to various downstream applications, such as dialog system \cite{madotto-etal-2018-mem2seq}, knowledge graph \cite{10.5555/2886521.2886624}, web search \cite{xiong2017explicit}, among others.

Many efforts have been devoted to improving the quality of extracted relation facts \cite{han-etal-2020-data}.
Conventional supervised relation extraction is oriented to \textbf{known} relations with pre-specified schema. Hence, the paradigm follows a \textit{closed-set setting}, meaning that during both training and testing the relations remain the same. Nowadays, neural RE methods have achieved remarkable success within this setting \cite{wang-etal-2016-relation,DBLP:journals/corr/abs-1905-08284}; 
and in contrast, open relation extraction (OpenRE) is focused on discovering constantly emerging \textbf{unknown} relations. Common practices include directly tagging the relational phrases that link entity pairs \cite{Zhan_Zhao_2020}, and clustering instances with the same relation \cite{hu-etal-2020-selfore,zhao-etal-2021-relation}.
However, relation extraction in real applications follows an \textit{open-set setting}, meaning that both known and unknown relations are \textbf{mixed} within testing data.\footnote{Some sentences even express no specific relations.} This requires that a model can not only distinguish among the known relations, but also filter the instances that express unknown relations. The ability to filter these instances is also called none-of-the-above (NOTA) detection \cite{gao-etal-2019-fewrel}.

Unfortunately, a well-performing closed-set model can still confidently make arbitrarily wrong predictions when exposed to unknown test data \cite{nguyen2015deep,recht2019imagenet}. As shown in fig. \ref{fig:intro} (a), the decision boundary is optimized only on the known relational data (white points), leading to a three-way partition of the whole space. Consequently, the unknown relational data (black points), especially those far from the decision boundary, will be confidently classified into one of the known relations. By contrast, a more compact decision boundary (as shown in fig. \ref{fig:intro} (b)) is desirable for NOTA detection. However, the compact decision boundary requires ``difficult'' negative data (red points in fig. \ref{fig:intro} (b)) to be used, so strong supervision signals can be provided. It is important to note that synthesizing such negative data is a non-trivial task.

In this work, we propose an unknown-aware training method, which simultaneously optimizes known relation classification and NOTA detection. To effectively regularize the classification, we iteratively generate negative instances and optimize a NOTA detection score. During the testing phase, instances with low scores are considered as NOTA and filtered out. 
The key of the method is to synthesize ``difficult'' negative instances. 
Inspired by text adversarial attacks, we achieve the goal by substituting a small number of critical tokens in original training instances. This would erase the original relational semantics and the model is not aware of it.
By using gradient-based token attribution and linguistic rules, key tokens that express the target relation are found. 
Then, the tokens are substituted by misleading normal tokens that would cause the greatest increase of NOTA detection score, thus misleading negative instances, which are more likely to be mistaken by the model as known relations, are synthesized.
Human evaluation shows that almost all the synthesized negative instances do not express any known relations. Experimental results show that the proposed method learns more compact decision boundary and achieve state-of-the-art NOTA detection performance. Our codes are publicly available at Github.\footnote{https://github.com/XinZhao0211/OpenSetRE.}

The contributions are threefold: (1) we propose a new unknown-aware training method for more realistic open-set relation extraction. The method achieves state-of-the-art NOTA detection, without compromising the classification of known relations; (2) the negative instances are more challenging to the model, when compared to the mainstream synthesis method \footnote{A quantitative analysis will be provided in Sec. \ref{sec:neg-ana}.} (e.g., generative adversarial network (GAN)-based method); (3) the comprehensive evaluation and analysis facilitate future research on the pressing but underexplored task.

\section{Related Works}
\textbf{Open-set Classification}:
The open-set setting considers knowledge acquired during training phase to be incomplete, thereby new unknown classes can be encountered during testing. The pioneering explorations in \cite{6365193} formalize the open-set classification task, and have inspired a number of subsequent works, which roughly fall into one of the following two groups. 

The first group explores model regularization using unknown data. \citet{larson-etal-2019-evaluation} manually collect unknown data to train a $(n+1)$-way classifier with one additional class, where $(n+1)^{th}$ class represents the unknown class. 
Instead of manually collecting unknown data, \citet{10.1109/TASLP.2020.2983593} generate feature vectors of unknown data using a generative adversarial network \cite{goodfellow2014generative}. \citet{zhan-etal-2021-scope} use MixUp technique \cite{NEURIPS2019_36ad8b5f} to synthesize known data into unknown data. 

The second group approaches this problem by discriminative representation learning, which facilitates open-set classification by widening the margin between known and unknown classes. MSP \cite{DBLP:conf/iclr/HendrycksG17} is a maximum posterior probability-based baseline and ODIN \cite{liang2018enhancing} enlarges the difference between known and unknown classes by adding temperature scaling and perturbations to MSP. More recently, different optimization objectives such as large margin loss \cite{lin-xu-2019-deep} and gaussian mixture loss \cite{yan-etal-2020-unknown} are adopted to learn more discriminative representations. \citet{shu-etal-2017-doc,xu-etal-2020-deep,zhang2021deep} also impose gaussian assumption to data distribution to facilitate distinct unknown data. 

    \begin{figure*}[t]
    \centering
        \includegraphics[width=\linewidth]{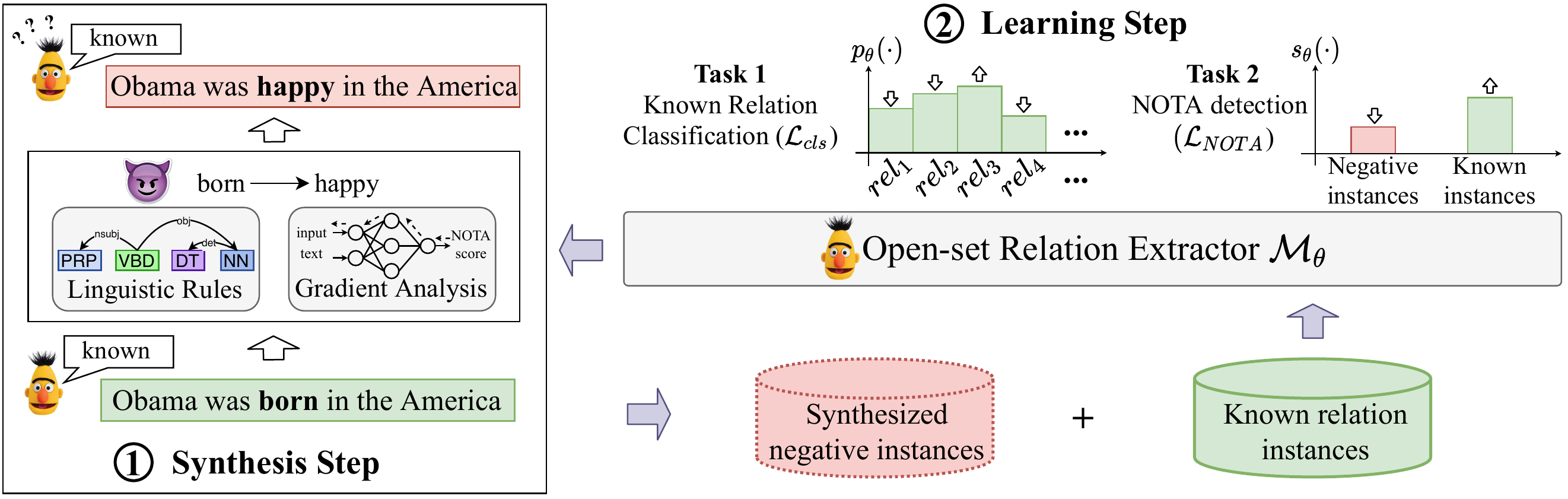}
        \caption{Overview of the proposed unknown-aware training method. The training loop consists of two iteration steps: the synthesis step consists in adaptively synthesizing ``difficult'' instances according to the states of the model; while in the learning step, an optimization of the dual objectives of both known relation classification and NOTA relation detection is performed, based on the known and synthesized instances.}
        \label{fig:model}
    \end{figure*}  
    
\noindent\textbf{Open-set Relation Extraction}: Open-set RE is a pressing but underexplored task. Most of the existing RE methods manually collect NOTA data and adopt a $(n+1)$ way classifier to deal with NOTA relations \cite{DBLP:journals/corr/abs-1809-10185,zhu-etal-2019-graph,ma-etal-2021-sent}. However, the collected NOTA data with manual bias cannot cover all NOTA relations and thus these methods cannot effectively deal with open-set RE \cite{gao-etal-2019-fewrel}. Our method avoids the bias and the expensive cost of manually collecting NOTA data by automatically synthesizing negative data. Compared with general open-set classification methods, our method takes relational linguistic rules into consideration and outperforms them by a large margin.

\section{Approach}
    We start by formulating the \textit{open-set} relation extraction task. Let $\mathcal{K}=\{r_1,...,r_n\}$ denote the set of known relations and \texttt{NOTA} indicates that the instance does not express any relation in $\mathcal{K}$. Given a training set $\mathcal{D}_{\text{train}}=\{(x_i,y_i)\}_{i=1}^N$ with $N$ positive samples, consisting of relation instance $x_i$ with a pre-specified entity pair \footnote{We assume that the entity recognition has already been done and an instance expresses at most one relation between the entity pair.} and relation $y_i\in\mathcal{K}$, we aim to learn a open-set relation extractor $\mathcal{M}=\{p_\theta(y|x), s_\theta(x)\}$, where $\theta$ denote the model parameters. $p_\theta(y|x)$ is the classification probability on the known relations (The \texttt{NOTA} label is excluded from $p_\theta(y|x)$). NOTA detection score $s_\theta(x)$ is used to distinguish between known relations and \texttt{NOTA}. $x$ is classified as \texttt{NOTA} if $s_\theta(x)$ is less than the threshold $\alpha$. Conversely, $x$ is classified into a known relation $\hat{y}=\mathop{\arg\max}_yp_\theta(y|x)$.
    
    \subsection{Method Overview}
    
    We approach the problem by an unknown-aware training method, which dynamically synthesizes ``difficult'' negative instances and optimizes the dual objectives of both known relation classification and NOTA detection.
    As shown in fig. \ref{fig:model}, the training loop consists of two iteration steps:

    \noindent\textbf{\ding{172} Synthesis Step}: This step aims to synthesize ``difficult'' negative instances for model regularization. We draw inspiration from text adversarial attacks to achieve the goal. Specifically, $\mathcal{B}=\{(x_i,y_i)\}_{i=1}^B$ represents a training batch sampled from $\mathcal{D}_{\text{train}}$. For each $(x,y)\in\mathcal{B}$, we synthesize a negative instance by substituting the key relational tokens of $x$ with misleading tokens. First, both the attribution method and relational linguistic rules are used to find key tokens expressing the target relation $y$. 
    Second, the misleading token $w_i^{\text{mis}}$ is searched for each key token $w_i$, along the direction of the gradient $\nabla_{\bm{w}_i}s_\theta(x)$. 
    By substituting $w_i$ with $w_i^{\text{mis}}$,
    it is expected for $s_\theta(x)$ to experience its greatest increase,
    so it is difficult for the model to correctly detect the derived negative instance $x^\prime$ as \texttt{NOTA}.

    \noindent\textbf{\ding{173} Learning Step}: This step aims to optimize the open-set relation extractor $\mathcal{M}=\{p_\theta(y|x), s_\theta(x)\}$. Based on the training batch $\mathcal{B}$ from $\mathcal{D}_{\text{train}}$, we optimize $p_\theta(y|x)$ to accurately classify known relations. To effectively detect \texttt{NOTA} instances, we further synthesize negative batch $\mathcal{B}^\prime=\{(x_i^\prime,\texttt{NOTA})\}_{i=1}^B$ and optimize the model to widen the gap of $s_\theta(x)$ between $x\in\mathcal{B}$ and $x^\prime\in\mathcal{B}^\prime$. Consequently, instances with low $s_\theta(x)$ scores are filtered out before being fed into $p_\theta(y|x)$.

    Next, we elaborate on the model structure of $\mathcal{M}$ (sec. \ref{sec:extractor}) and the technical details of the synthesis step (sec. \ref{sec:neg}) and the learning step (sec. \ref{sec:obj}).
    
\subsection{Open-set Relation Extractor}
\label{sec:extractor}
\textbf{Instance Encoder and Classifier}: Given an input instance $x=\{w_1,..,w_n\}$ with four reserved special tokens $[E_1],[\backslash E_1],[E_2],[\backslash E_2]$ marking the beginning and end of the head and tail entities, the instance encoder aims to encode the relational semantics into a fixed-length representation $\bm{h}=\bm{enc}(x)\in \mathbb{R}^d$. We adopt BERT \cite{DBLP:journals/corr/abs-1810-04805}, a common practice, as the implementation of the encoder. We follow \citet{baldini-soares-etal-2019-matching} to concatenate the hidden states of special tokens $[E_1]$ and $[E_2]$ as the representation of the input instance.
        \begin{gather}
            \bm{w}_1,..,\bm{w}_n={\rm BERT}(w_1,..,w_n)\\
            \bm{h}=\bm{w}_{[E_1]}\oplus\bm{w}_{[E_2]},
        \end{gather}
where $\bm{w}_i, \bm{w}_{[E_1]}, \bm{w}_{[E_2]}$ denotes the hidden states of token $w_i, [E_1], [E_2]$, respectively. $\oplus$ denotes the concatenation operator. The classification probability on known relations $p_\theta(\cdot|x)$ can be derived through a linear head $\bm{\eta}(\cdot)$:
        \begin{gather}
            \bm{\eta}(\bm{h})=W_{\text{cls}}\bm{h}+b\\
            p_\theta(\cdot|x)=\text{Softmax}(\bm{\eta}(\bm{h})),
            \label{eq:cls}
        \end{gather}
where $W_{\text{cls}}\in\mathbb{R}^{n\times d}$ is the weight matrix transforming the relation representation to the logits on $n$ known relations and $b$ is the bias.

\noindent\textbf{NOTA Detection Score}: The goal of distinguishing between known and NOTA relations requires the modeling of the data density. However, directly estimating $\log p(x)$ can be computationally intractable because it requires sampling from the entire input space. Inspired by \citet{WeitangLiu2020EnergybasedOD} in the image understanding task, the free energy function $E(\bm{h})$ is theoretically proportional to the probability density of training data. Considering that it can be easily derived from the linear head $\bm{\eta}(\cdot)$ without additional calculation, the negative free energy function is used to compute the NOTA detection score as follows:
        \begin{equation}
            s_\theta(x)=-E(\bm{h})=\log\sum_{j=1}^ne^{\bm{\eta}(\bm{h})_j},
        \end{equation}
where $\bm{\eta}(\bm{h})_j$ denotes the $j^{\text{th}}$ logit value of $\bm{\eta}(\bm{h})$. The detection score has shown to be effective in out-of-distribution detection \cite{WeitangLiu2020EnergybasedOD}. Based on the classification probability $p_\theta(\cdot|x)$ and NOTA detection score $s_\theta(x)$, the open-set relation extractor $\mathcal{M}$ works in the following way:
\begin{gather}
\hat{y}=\left\{
    \begin{array}{cl}
    \arg\max_yp_\theta(y|x)             & {\mathcal{S}(x)>\alpha}\\
    \texttt{NOTA}          & {\mathcal{S}(x)\leq\alpha},
    \end{array} \right.
    \end{gather}
where $\alpha$ is the detection threshold.

\subsection{Iterative Negative Instances Synthesis}
\label{sec:neg}

``Difficult'' negative instances are the key to effective model regularization. $x=\{w_1,..,w_n\}$ is a training instance with a label $y$. To synthesize negative instance $x^\prime$, we perturb each key token $w_i$, which expresses the relation $y$, with a misleading token $w_i^{\text{mis}}$. The substitutions are expected to erase original relational semantics without the model being aware of it. Based on the attribution technique and relational linguistic rules, a score $I(w_i,x,y)$ is developed to measure the contribution of a token $w_i \in x$ to relation $y$ as follows:
\begin{equation}
    I(w_i,x,y)=a(w_i,x)\cdot t(w_i,y)\cdot dp(w_i,x),
\end{equation}
where $a(w_i,x)$ denotes an attribution score reweighted by two linguistic scores $t(w_i,y)$, $dp(w_i,x)$. We rank all tokens according to $I(w_i,x,y)$ in descending order and take the first $\epsilon$ percent of tokens as key tokens to perform substitutions. Next, we elaborate on (1) how to calculate the attribution score $a(w_i,x)$ and linguistic scores $t(w_i,y)$, $dp(w_i,x)$; (2) how to select misleading tokens for substitution.

\noindent\textbf{Gradient-based Token Attribution}:
Ideally, when the key tokens are removed, instance $x$ will no longer express the original known relation $y$, and the NOTA detection score $s_\theta(x)$ would drop accordingly. Therefore, the contribution of a token $w_i$ to relational semantics can be measured by a counterfactual:
    \begin{equation}
        c(w_i,x)=s_\theta(x)-s_\theta(x_{-w_i}),    
    \end{equation}
where $x_{-w_i}$ is the instance after removing $w_i$. However, to calculate the contribution of each token in instance $x$, $n$ forward passes are needed, which is highly inefficient. Fortunately, a first-order approximation of contribution $c(w_i,x)$ can be obtained by calculating the dot product of word embedding $\bm{w}_i$ and the gradient of $s_\theta(x)$ with respect to $\bm{w}_i$, that is $\nabla_{\bm{w}_i}s_\theta(x)\cdot \bm{w}_i$ \cite{feng-etal-2018-pathologies}. The contribution of $n$ tokens can thus be computed with a single forward-backward pass. Finally, a normalized attribution score is used, in order to represent the contribution of each token:
\begin{equation}
    a(w_i,x)=\frac{|\nabla_{\bm{w}_i}s_\theta(x)\cdot \bm{w}_i|}{\sum_{j=1}^n|\nabla_{\bm{w}_j}s_\theta(x)\cdot \bm{w}_j|}.
\end{equation}

\noindent\textbf{Linguistic Rule-based Token Reweighting}: As a supplement to the attribution method, linguistic rules that describe the pattern of relational phrases can provide valuable prior knowledge for the measure of tokens' contribution. Specifically, the following two rules are used. Rule $1$: \textit{If a token $w_i$ significantly contributes to relation $y$, it should appear more frequently in the instances of $y$, and rarely in the instances of other relations}. By following this rule, \texttt{tf-idf} statistic \cite{10.5555/866292} $t(w_i,y)$\footnote{The statistic is based on the whole training set and does not change with a specific instance $x$.} is used to reflect the contribution of token $w_i$ to relation $y$ (Appendix \ref{sec:app-tf} contains additional details about the statistic). Rule $2$: \textit{Tokens that are part of the dependency path between the entity pair usually express the relation between the entity pair, while shorter dependency paths are more likely to represent the relation} \cite{DBLP:journals/corr/abs-1801-07174}. Following the rule, stanza\footnote{https://stanfordnlp.github.io/stanza/depparse.html} is used to parse the instance and the dependency score as calculated as follows:
\begin{gather}
    \label{eq:dp}
    dp(w_i,x)=\left\{
    \begin{array}{cc}
     |x|/|\mathcal{T}|& {w_i\in \mathcal{T}}\\
    1,& {otherwise},
    \end{array} \right.
\end{gather}
where $\mathcal{T}$ denotes the set of tokens in the dependency path between the entity pair. $|x|$, $|\mathcal{T}|$ denote the number of tokens in instance $x$ and set $\mathcal{T}$, respectively.
Eq. \ref{eq:dp} indicates that the tokens in $\mathcal{T}$ are given a higher weight, and the shorter the path, the higher the weight.

\noindent\textbf{Misleading Token Selection}: Negative instances are synthesized by substituting key tokens with misleading tokens. Note that we have obtained the gradient of $s_\theta(x)$ with respect to each token $w_i$ in the attribution step. Based on the gradient vectors, a misleading token is selected from vocabulary $\mathcal{V}$ for each key token $w_i$ as follows:
\begin{equation}
    w_i^{\text{mis}}=\mathop{\arg\max}_{w_j\in \mathcal{V}}\nabla_{\bm{w}_i}s_\theta(x)\cdot \bm{w}_j.
\end{equation}
Substituting $w_i$ with $w_i^{\text{mis}}$ is expected to cause the greatest increase in $s_\theta(x)$, so the synthesized negative instance is misleading to the model.
To avoid that $w_i^{mis}$ is also a key token of a known relation, the top 100 tokens with the highest \texttt{tf-idf} statistic of each relation are removed from the vocabulary $\mathcal{V}$, when performing the substitution. Human evaluation results show that almost all the synthesized negative instances do not express any known relation. In addition, we provide two real substitution cases in tab. \ref{tab:case}.

\subsection{Unknown-Aware Training Objective}
\label{sec:obj}
In this section, we introduce the unknown-aware training objective for open-set relation extraction. Based on the synthesized negative samples, an optimization of the dual objectives of both known relation classification and NOTA relation detection is performed. 
Specifically, at the $m^{\text{th}}$ training step, A batch of training data $\mathcal{B}_m=\{(x_i,y_i)\}_{i=1}^{B}$ is sampled from $\mathcal{D}_{\text{train}}$. Cross entropy loss is used for the optimization of known relation classification: 
\begin{equation}
    \mathcal{L}_{cls}=\frac{1}{B}\sum_{i=1}^{B}(-\log p_\theta(y_i|x_i)),
\end{equation}
where $p_\theta(\cdot|x_i)$ is the classification probability on the known relations (eq. \ref{eq:cls}). For each instance $x$ in $\mathcal{B}_m$, we synthesize a negative sample $x^\prime$ as described in sec. \ref{sec:neg}, and finally obtain a batch of negative samples $\mathcal{B}_m^\prime=\{(x_i^\prime,\texttt{NOTA})\}_{i=1}^B$.  
To learn a compact decision boundary for NOTA detection, we use the binary sigmoid loss to enlarge the gap of detection scores $s_\theta(\cdot)$ between known and synthesized instances as follows:
\begin{gather}
    \begin{split}
        \mathcal{L}_{\text{NOTA}}=-\frac{1}{B}\sum_{i=1}^{B}\log\sigma(s_\theta(x_i))\\
        -\frac{1}{B}\sum_{i=1}^{B}\log(1-\sigma(s_\theta(x_i^\prime)))
    \end{split}
\end{gather}
where $\sigma(x)=\frac{1}{1+e^{-x}}$ is the \texttt{sigmoid} function. The overall optimization objective is as follows:
\begin{gather}
    \mathcal{L}=\mathcal{L}_{cls}+\beta\cdot\mathcal{L}_{\text{NOTA}},
\end{gather}
where $\beta$ is a hyper-parameter to balance the two loss term. 

        \begin{table*}
            \centering
            \resizebox{\linewidth}{!}{
            \begin{tabular}{l c ccc cc}
            \toprule
            \multirow{2}{*}{\textbf{Method}} & \multicolumn{3}{c}{\textbf{FewRel}} & \multicolumn{3}{c}{\textbf{TACRED}}\\
            \cmidrule(r){2-4}\cmidrule(r){5-7}
            %\cmidrule(r){3-4}\cmidrule(r){5-6}\cmidrule(r){7-8}\cmidrule(r){9-10}
            & ACC$\uparrow$ & AUROC$\uparrow$ & FPR95$\downarrow$ &  ACC$\uparrow$ & AUROC$\uparrow$ & FPR95$\downarrow$\\
            \midrule
            MSP \citep{DBLP:conf/iclr/HendrycksG17} & 63.69$_{1.71}$ & 83.60$_{2.12}$ & 62.93$_{4.05}$ & 71.83$_{1.99}$ & 89.24$_{0.32}$ & 43.20$_{4.15}$ \\
            DOC \cite{shu-etal-2017-doc} &63.96$_{1.00}$ & 84.46$_{0.97}$ & 59.38$_{1.92}$ & 70.08$_{0.59}$ & 89.40$_{0.25}$ & 42.83$_{1.66}$ \\
            ODIN \cite{liang2018enhancing} & 66.78$_{1.57}$ & 84.47$_{2.16}$ & 55.98$_{3.03}$ & 72.37$_{2.32}$ & 89.42$_{0.30}$ & 40.83$_{3.09}$  \\
            MixUp \cite{thulasidasan2019mixup} & 66.30$_{0.45}$ & 84.95$_{1.38}$ & 57.44$_{0.37}$ & 72.85$_{1.60}$ & 89.80$_{0.59}$ & 40.30$_{3.77}$\\
            Energy \cite{WeitangLiu2020EnergybasedOD} & 71.54$_{1.05}$ & 85.53$_{1.84}$ & 46.88$_{1.50}$ & 75.15$_{0.14}$ & 90.34$_{0.12}$ & 35.30$_{2.86}$ \\
            Convex \cite{zhan-etal-2021-scope} & 71.19$_{1.51}$ & 86.23$_{0.81}$ & 46.00$_{2.67}$ & 71.55$_{1.17}$ & 90.16$_{0.58}$ & 37.40$_{3.28}$ \\
            SCL \cite{zeng-etal-2021-modeling} & 65.52$_{1.48}$ & 86.71$_{1.23}$ & 58.04$_{3.24}$ & 72.70$_{2.17}$ & 90.22$_{0.67}$ & 35.80$_{3.67}$ \\
            \textbf{Ours}&\textbf{74.00}$_{0.56}$ & \textbf{88.73}$_{0.67}$ & \textbf{41.17}$_{1.37}$ & \textbf{76.97}$_{1.81}$ & \textbf{91.02}$_{0.59}$ & \textbf{30.27}$_{2.29}$ \\
            \bottomrule
            \end{tabular}
            }
            \caption{Main results of open-set relation extraction. The subscript represents the corresponding standard deviation (e.g., 74.00$_{0.56}$ indicates 74.00$\pm0.56$). The results of \textbf{ACC} on $n$ known relations are provided in tab.\ref{tab:known}.} 
            \label{tab:main_res}
        \end{table*}       

\section{Experimental Setup}
%In this section, we first describe the datasets used to evaluate the proposed method. Then, we detail the baseline models for comparison and introduce the evaluation metrics. Finally, we clarify the implementation details of the proposed method.
\subsection{Datasets}
%We conduct our experiments on two well-known relation extraction datasets including FewRel and TACRED.

\noindent\textbf{FewRel} \cite{han-etal-2018-fewrel}. FewRel is a human-annotated dataset, which contains 80 types of relations, each with 700 instances. We take the top 40 relations as known relations. The middle 20 relations are taken as unknown relations for validation. And the remaining 20 relations are unknown relations for testing. Our training set contains 22,400 instances from the 40 known relations. Both the validation and test set consist of 5,600 instances, of which 50$\%$ are from unknown relations. \textbf{Note that the unknown relations in the test set and the validation set do not overlap.}

\noindent\textbf{TACRED} \cite{zhang-etal-2017-position}. TACRED is a large-scale relation extraction dataset, which contains 41 relations and a \texttt{no\_relation} label indicating no defined relation exists. Similar to FewRel, we take the top 21 relations as known relations. The middle 10 relations are taken as unknown relations for validation. The remaining 10 relations and \texttt{no\_relation} are unknown relations for testing. We randomly sample 9,784 instances of known relations to form the training set. Both the validation and test set consist of 2,000 instances, of which 50$\%$ are from unknown relations. Unknown relations in the validation set and the test set still do not overlap.

For the specific composition of relations in each dataset, please refer to Appendix \ref{sec:rel}.

\subsection{Compared Methods}
To evaluate the effectiveness of the proposed method, we compare our method with mainstream open-set classification methods, which can be roughly grouped into the following categories: \textbf{MSP} \cite{DBLP:conf/iclr/HendrycksG17}, \textbf{DOC} \cite{shu-etal-2017-doc}, \textbf{ODIN} \cite{liang2018enhancing}, \textbf{Energy} \cite{WeitangLiu2020EnergybasedOD}, and \textbf{SCL} \cite{zeng-etal-2021-modeling} detect unknown data through a carefully designed score function or learning a more discriminative representation. No synthesized negative instances are used in these methods. \textbf{MixUp} \cite{thulasidasan2019mixup}, and \textbf{Convex} \cite{zhan-etal-2021-scope} use synthesized negative instances to regularize the model. Please refer to the appendix \ref{sec:app-baseline} for a brief introduction to these methods.

We do not compare \textbf{BERT-PAIR} \cite{gao-etal-2019-fewrel} because it is only applicable to the few-shot setting. We use \textbf{DOC} \cite{shu-etal-2017-doc} with a BERT encoder as an alternative method for it.

\subsection{Metrics}
Following previous works \cite{WeitangLiu2020EnergybasedOD,zeng-etal-2021-modeling}, we treat all unknown instances as one \texttt{NOTA} class and adopt three widely used metrics for evaluation.
(1) \textbf{FPR95}: The false positive rate of \texttt{NOTA} instances when the true positive rate of known instances is at 95\%. The smaller the value, the better. 
(2) \textbf{AUROC}: the area under the receiver operating characteristic curve. It is a threshold-free metric that measures how well the detection score ranks the instances of known and \texttt{NOTA} relations. (3) \textbf{ACC}: The classification accuracy on $n$ known relations and one \texttt{NOTA} relation, measuring the overall performance of open-set RE.

\subsection{Implementation Details}
	We use the AdamW as the optimizer, with a learning rate of $2e-5$ and batch size of $16$ for both datasets. Major hyperparameters are selected with grid search according to the model performance on a validation set. The detection threshold is set to the value at which the true positive rate of known instances is at 95\%. The regularization weight $\beta$ is 0.05 selected from $\{0.01, 0.05, 0.1, 0.15, 0.5\}$. See the appendix \ref{sec:app-sub} for the processing of sub-tokens. The dependency parsing is performed with stanza 1.4.2. All experiments are conducted with Python 3.8.5 and PyTorch 1.7.0, using a GeForce GTX 2080Ti with 12GB memory. 
        \begin{figure}[t]
            \includegraphics[width=\columnwidth]{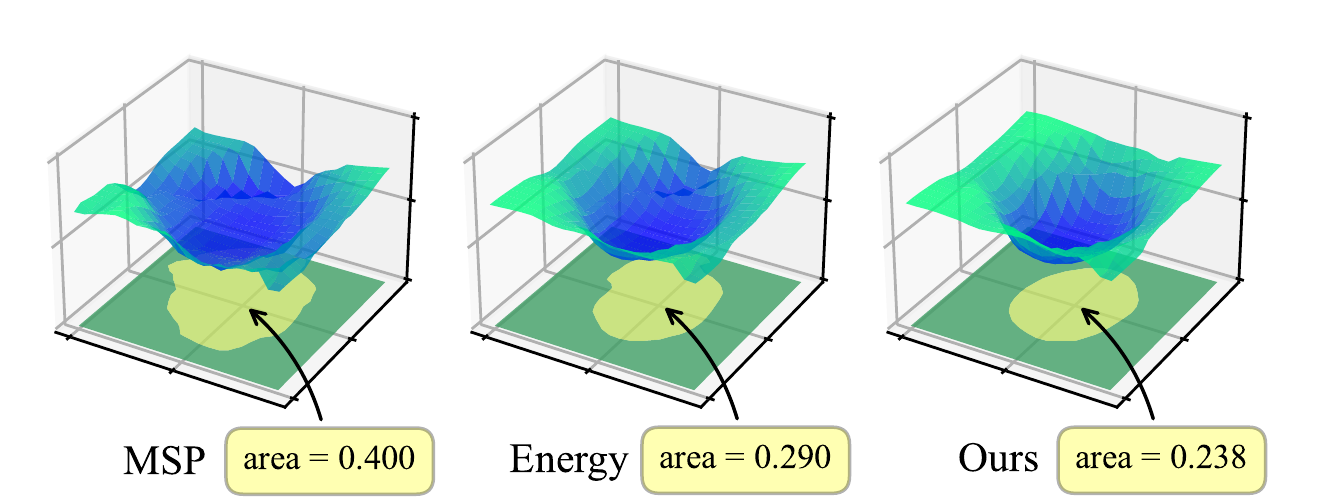}
            \caption{Decision boundary visualization. Energy can be seen as a degenerate version of our method when removing unknown-aware training. The vertical axis represents the difference between the detection threshold $\alpha$ and the \texttt{NOTA} score $s_\theta(x)$, normalized to the range of $[-1, 1]$. When an instance falls within the yellow region below zero, the model classifies it as a known relation. Conversely, when a sentence falls within the green region above zero, the model identifies it as \texttt{NOTA}.} 
            \label{fig:vis}
        \end{figure}
        
        \begin{table*}[t]
            \centering
            \resizebox{\linewidth}{!}{
            \begin{tabular}{l cccc cccc}
            \toprule
            \multirow{2}{*}{\textbf{Method}} & \multicolumn{4}{c}{\textbf{FewRel}} & \multicolumn{4}{c}{\textbf{TACRED}}\\
            \cmidrule(r){2-5}\cmidrule(r){6-9}
            %\cmidrule(r){3-4}\cmidrule(r){5-6}\cmidrule(r){7-8}
            &ACC$\uparrow$ & AUROC$\uparrow$ &FPR95$\downarrow$& $\Delta s_\theta\downarrow$ &ACC$\uparrow$ & AUROC$\uparrow$ &FPR95$\downarrow$& $\Delta s_\theta\downarrow$\\
            \midrule
            
            Baseline & 71.54 & 85.53 & 46.88 & $-$ & 75.15 & 90.34 & 35.30 & $-$\\
            Gaussian  & 71.81 & 86.67 & 46.81 & 4.35 & 74.73 & 90.16 & 35.47 & 4.48\\
            Gaussian$^\dag$ & \underline{72.93} & 86.66 & \underline{42.69} & \textbf{0.02} & 75.17 & 90.38 & 34.73 & \textbf{0.03}\\
            MixUp& 72.86 & 86.17 & 43.90 & 2.34 & 75.95 & 89.35 & \underline{33.20} & 1.90\\
            Real& 71.75 & 86.52 & 46.08 & 3.55 & \underline{76.10} & 89.92 & 33.67 & 3.91\\
            GAN& 72.11 & \underline{86.77} & 45.69 & 4.01 & 76.06 & \underline{90.46} & 34.30 & 4.10\\
            \textbf{Ours} & \textbf{74.00} & \textbf{88.73} & \textbf{41.17} & \underline{1.73} & \textbf{76.97} & \textbf{91.02}&\textbf{30.27}&\underline{1.36}\\
            \bottomrule
            \end{tabular}
            }
            \caption{The unknown-aware training with various negative instance synthesis methods. The numbers in \textbf{bold} and \underline{underlined} indicate the best and second-best results, respectively. To quantify the difficulty of negative instances, we calculate the average difference $\Delta s_\theta$ between the NOTA detection score of known and negative instances. Obviously, the smaller the difference, the more difficult it is for the model to distinguish the two types of instances.} 
            \label{tab:syn}
        \end{table*}     
        
\section{Results and Analysis}    
\subsection{Main Results}
In this section, we evaluate the proposed method by comparing it with several competitive open-set classification methods. The results are reported in tab. \ref{tab:main_res}, from which we can observe that our method achieves state-of-the-art NOTA detection (reflected by FPR95 and AUROC) without compromising the classification of known relations (reflected by ACC). In some baseline methods (e.g., MSP, ODIN, Energy, SCL), only instances of known relations are used for training. Compared with them, we explicitly synthesize the negative instances to complete the missing supervision signals, and the improvement in NOTA detection shows the effectiveness of the unknown-aware training. To intuitively show the changes of the decision boundary, we use the method of \citet{ijcai2019p0583} to visualize the decision boundary of the model in the input space. As can be seen from fig. \ref{fig:vis}, a more compact decision boundary is learned with the help of unknown-aware training. Although methods such as MixUp, and Convex also synthesized negative instances, our method is still superior to them. This may be due to the fact that our negative instances are more difficult and thus beneficial for an effective model regularization (we provide more results in sec. \ref{sec:neg-ana} to support the claim).

\subsection{Negative Instance Synthesis Analysis}
\label{sec:neg-ana}
%The synthesis of high-quality negative instances is non-trivial. 
In this section, the unknown-aware training objective is combined with the various negative instance synthesis methods to fairly compare the performance of these synthesis methods. The results are shown in tab. \ref{tab:syn}. \texttt{Baseline} means no negative instances are used. \texttt{Gaussian} takes Gaussian noise as negative instances and \texttt{Gaussian}$^\dag$ adds the noise to known instances. \texttt{MixUp} synthesizes negative instances by convexly combining pairs of known instances. \texttt{Real} means using real NOTA instances\footnote{We use the data from SemEval-2010 \cite{hendrickx-etal-2010-semeval}. The overlap relations are manually removed.}. \texttt{GAN} synthesizes negative instances by Generative Adversarial Network \cite{ryu-etal-2018-domain}. 

\noindent\textbf{Correlation between effectiveness and difficulty}. 
(1) \texttt{Gaussian} with the largest $\Delta s_\theta$ performs even worse than \texttt{Baseline} in TACRED, suggesting that overly simple negative instances are almost ineffective for model regularization. 
(2) Our method synthesizes the second difficult negative instances (reflected by $\Delta s_\theta$) and achieves the best performance (reflected by ACC, AUROC, FPR95), which shows that the difficult negative instances are very beneficial for effective model regularization. (3) The difficulty of negative instances of competitive methods (e.g., \texttt{MixUp}, \texttt{Real}, \texttt{GAN}) is lower than that of \texttt{Ours}, which indicates that it is non-trivial to achieve our difficulty level.
(4) Although \texttt{Gaussian}$^\dag$ synthesizes the most difficult negative instances, our method still significantly outperforms \texttt{Gaussian}$^\dag$. One possible reason is that overly difficult instances may express the semantics of known relations. This leads to the following research question.

\noindent\textbf{Do our synthetic negative instances really not express any known relations}?
        \begin{table}
            \centering
            \resizebox{\columnwidth}{!}{
            \begin{tabular}{l p{1.8cm}<{\centering}p{1.8cm}<{\centering}p{1.8cm}<{\centering}p{1.8cm}<{\centering}}
            \toprule
            \textbf{Dataset}& NOTA & \makecell[c]{Known-\\Original} & \makecell[c]{Known-\\Other} & Controversial\\
            \midrule
            \textbf{FewRel}& 92 & 2 & 1 & 5\\
            \textbf{TACRED}& 90 & 3 & 0 & 7\\
            \bottomrule
            \end{tabular}
            }
            \caption{Human evaluation of our negative instances. More than 90\% of the negative instances do not express any known relations.}
            \label{tab:human}
        \end{table}  
We conduct human evaluation to answer this question. Specifically, we randomly select 100 synthesized negative instances on each dataset and asked human judges whether these instances express known or \texttt{NOTA} relations. The evaluation is completed by three independent human judges. We recruit 3 graduates in computer science and English majors from top universities.
All of them passed a test batch. Each graduate is paid \$8 per hour. The results are shown in tab. \ref{tab:human}, from which we can observe that: (1) More than 90\% of the negative instances do not express any known relations (\texttt{NOTA}). (2) Very few instances remain in the original known relations (\texttt{Known-Original}) or are transferred to another known relation (\texttt{Known-Other}). (3) There are also some instances that are \texttt{Controversial}. Some volunteers believe that the instances express known relations, while others believe that the instances are \texttt{NOTA}. In general, our synthesis method achieves satisfactory results, but there is still potential for further improvement.

\subsection{Ablation Study}

        \begin{table}
            \centering
            \resizebox{\columnwidth}{!}{
            \begin{tabular}{l cccc}
            \toprule
            \textbf{Method}& ACC$\uparrow$ & AUROC$\uparrow$ &FPR95$\downarrow$\\
            \midrule
            w/o attribution score& 73.81 & 88.34 & 41.32\\
            w/o dependency score & 73.89 & 88.55 & 41.88\\
            w/o tfidf statistic& 73.92 & 87.64 & 42.42\\
            w/o iterative synthesis & 72.61 & 86.90 & 44.71\\
            w/o misleading tokens & 71.87 & 86.99 & 46.35\\
            \textbf{Ours} & \textbf{74.00} & \textbf{88.73} & \textbf{41.17}\\
            \hline\hline
            w/o attribution score& 75.47 & 90.71 & 35.10\\
            w/o dependency score & 76.73 & 90.93 & 30.57\\
            w/o tfidf statistic & 76.68 & 90.46 & 34.43\\
            w/o iterative synthesis & 76.75 & 90.57 & 32.77\\
            w/o misleading tokens & 75.80 & 90.41 & 33.53\\
            \textbf{Ours}& \textbf{76.97} & \textbf{91.02} & \textbf{30.27}\\
            \bottomrule
            \end{tabular}
            }
            \caption{Abalation study of our method. The upper (resp. lower) part lists the results on FewRel (resp. TACRED).}
            \label{tab:abalation}
        \end{table}     
To study the contribution of each component in our method, we conduct ablation experiments on the two datasets and show the results in tab. \ref{tab:abalation}. First, the attribution score measures the impact of a token on NOTA detection of the model. The dependency score and \texttt{tf-idf} statistic reflect the matching degree between a token and the relational linguistic rules. When the three scores are removed, there may be some key relational phrases that can not be correctly identified and the performance decline accordingly. It is worth mentioning that the model parameters change dynamically with the training process, thus iteratively synthesizing negative instances is crucial for effective regularization. When the practice is removed, the static negative instances can not reflect the latest state of the model, and thus the performance degrades significantly. Finally, we remove misleading token selection by substituting the identified key tokens with a special token \texttt{[MASK]} and the performance is seriously hurt, which indicates that misleading tokens play an important role in synthesizing difficult instances.

\subsection{Hyper-parameter Analysis}
        \begin{figure}[t]
            \includegraphics[width=\columnwidth]{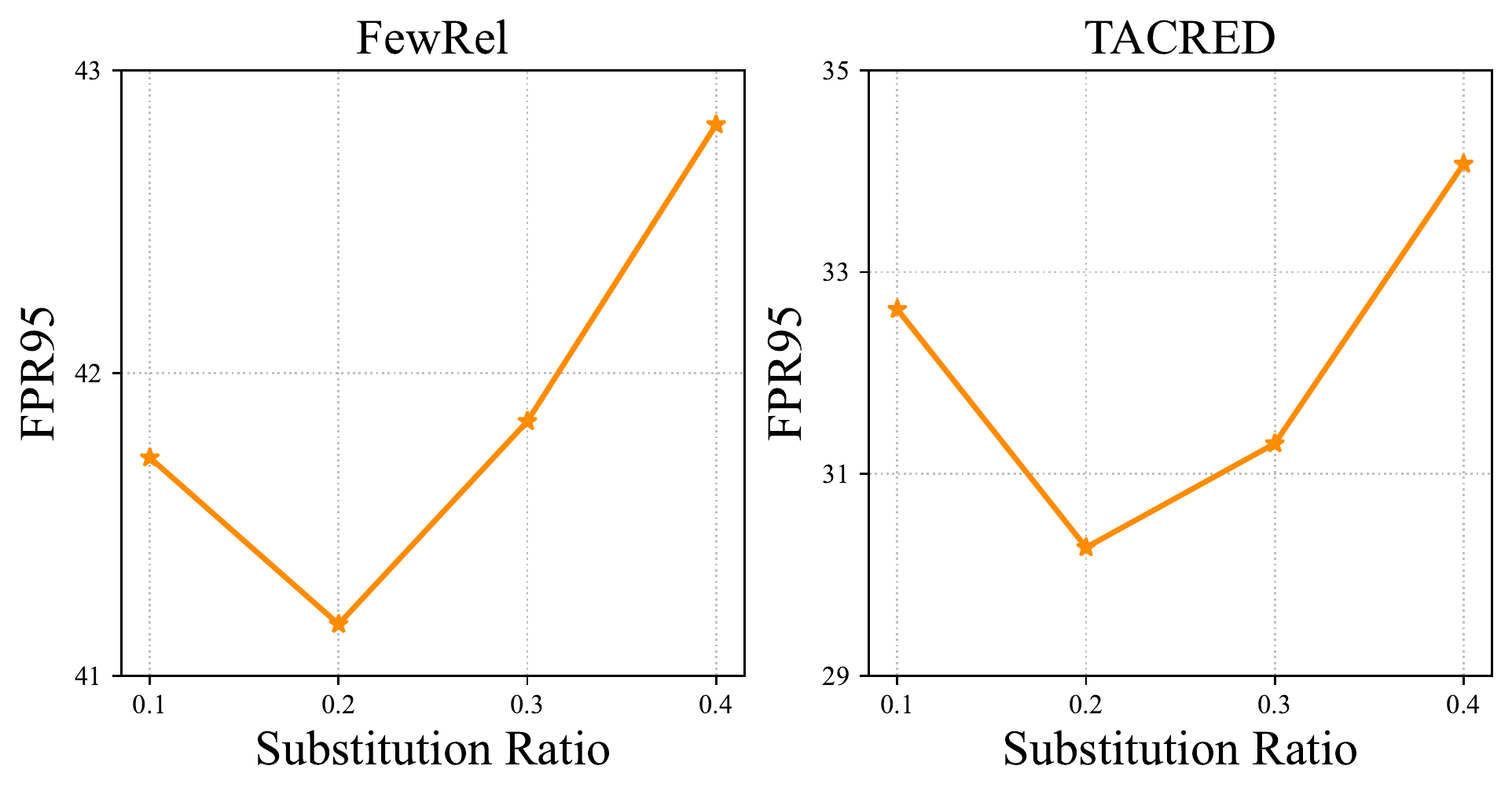}
            \caption{FPR95 with different substitution ratio.}
            \label{fig:ratio}
        \end{figure}
We synthesize negative instances by substituting $\epsilon$ percent of key tokens with misleading tokens. In this section, we conduct experiments to study the influence of substitution ratio $\epsilon$ on NOTA detection. From fig. \ref{fig:ratio} we obtain the following observations. When the substitution ratio gradually increases from 0, the performance of NOTA detection is also improved (Note that the smaller the value of FPR95, the better). This means that an overly small substitution ratio is not sufficient to remove all relational phrases. The residual relational tokens are detrimental to model regularization. When the substitution ratio exceeds a certain threshold (i.e., 0.2), a continued increase in the substitution ratio will lead to a decline in detection performance. One possible reason is that too high a substitution ratio can severely damage the original sentence structure, resulting in negative instances that differ too much from the real NOTA instances.

\section{Conclusions}
In this work, we propose an unknown-aware training method for open-set relation extraction, which is a pressing but underexplored task. We dynamically synthesize negative instances by the attribution technique and relational linguistic rules to complete the missing supervision signals. The negative instances are more difficult than that of other competitive methods and achieve effective model regularization. Experimental results show that our method achieves state-of-the-art NOTA detection without compromising the classification of known relations. We hope our method and analysis can inspire future research on this task.

\section*{Limitations}
We synthesize negative instances by substituting relational phrases with misleading tokens. However, the relational semantics in some instances may be expressed implicitly. That is, there are no key tokens that directly correspond to the target relation. Therefore, we cannot synthesize negative instances based on these instances. Additionally, we consider substitution ratio $\epsilon$ as a fixed hyperparameter. It may be a better choice to dynamically determine $\epsilon$ based on the input instance. We leave these limitations as our future work.

\section*{Acknowledgements}
The authors wish to thank the anonymous reviewers for their helpful comments. This work was partially funded by National Natural Science Foundation of China (No.62206057,62076069,61976056), Shanghai Rising-Star Program (23QA1400200), Program of Shanghai Academic Research Leader under grant 22XD1401100, and Natural Science Foundation of Shanghai (23ZR1403500).

\bibliography{anthology,custom}

\begin{thebibliography}{41}
\expandafter\ifx\csname natexlab\endcsname\relax\def\natexlab#1{#1}\fi

\bibitem[{Baldini~Soares et~al.(2019)Baldini~Soares, FitzGerald, Ling, and
  Kwiatkowski}]{baldini-soares-etal-2019-matching}
Livio Baldini~Soares, Nicholas FitzGerald, Jeffrey Ling, and Tom Kwiatkowski.
  2019.
\newblock \href {https://doi.org/10.18653/v1/P19-1279} {Matching the blanks:
  Distributional similarity for relation learning}.
\newblock In \emph{Proceedings of the 57th Annual Meeting of the Association
  for Computational Linguistics}, pages 2895--2905, Florence, Italy.
  Association for Computational Linguistics.

\bibitem[{Devlin et~al.(2018)Devlin, Chang, Lee, and
  Toutanova}]{DBLP:journals/corr/abs-1810-04805}
Jacob Devlin, Ming{-}Wei Chang, Kenton Lee, and Kristina Toutanova. 2018.
\newblock \href {http://arxiv.org/abs/1810.04805} {{BERT:} pre-training of deep
  bidirectional transformers for language understanding}.
\newblock \emph{CoRR}, abs/1810.04805.

\bibitem[{ElSahar et~al.(2018)ElSahar, Demidova, Gottschalk, Gravier, and
  Laforest}]{DBLP:journals/corr/abs-1801-07174}
Hady ElSahar, Elena Demidova, Simon Gottschalk, Christophe Gravier, and
  Fr{\'{e}}d{\'{e}}rique Laforest. 2018.
\newblock \href {http://arxiv.org/abs/1801.07174} {Unsupervised open relation
  extraction}.
\newblock \emph{CoRR}, abs/1801.07174.

\bibitem[{Feng et~al.(2018)Feng, Wallace, Grissom~II, Iyyer, Rodriguez, and
  Boyd-Graber}]{feng-etal-2018-pathologies}
Shi Feng, Eric Wallace, Alvin Grissom~II, Mohit Iyyer, Pedro Rodriguez, and
  Jordan Boyd-Graber. 2018.
\newblock \href {https://doi.org/10.18653/v1/D18-1407} {Pathologies of neural
  models make interpretations difficult}.
\newblock In \emph{Proceedings of the 2018 Conference on Empirical Methods in
  Natural Language Processing}, pages 3719--3728, Brussels, Belgium.
  Association for Computational Linguistics.

\bibitem[{Gao et~al.(2019)Gao, Han, Zhu, Liu, Li, Sun, and
  Zhou}]{gao-etal-2019-fewrel}
Tianyu Gao, Xu~Han, Hao Zhu, Zhiyuan Liu, Peng Li, Maosong Sun, and Jie Zhou.
  2019.
\newblock \href {https://doi.org/10.18653/v1/D19-1649} {{F}ew{R}el 2.0: Towards
  more challenging few-shot relation classification}.
\newblock In \emph{Proceedings of the 2019 Conference on Empirical Methods in
  Natural Language Processing and the 9th International Joint Conference on
  Natural Language Processing (EMNLP-IJCNLP)}, pages 6250--6255, Hong Kong,
  China. Association for Computational Linguistics.

\bibitem[{Goodfellow et~al.(2014)Goodfellow, Pouget-Abadie, Mirza, Xu,
  Warde-Farley, Ozair, Courville, and Bengio}]{goodfellow2014generative}
Ian Goodfellow, Jean Pouget-Abadie, Mehdi Mirza, Bing Xu, David Warde-Farley,
  Sherjil Ozair, Aaron Courville, and Yoshua Bengio. 2014.
\newblock Generative adversarial nets.
\newblock \emph{Advances in neural information processing systems}, 27.

\bibitem[{Han et~al.(2020)Han, Gao, Lin, Peng, Yang, Xiao, Liu, Li, Zhou, and
  Sun}]{han-etal-2020-data}
Xu~Han, Tianyu Gao, Yankai Lin, Hao Peng, Yaoliang Yang, Chaojun Xiao, Zhiyuan
  Liu, Peng Li, Jie Zhou, and Maosong Sun. 2020.
\newblock \href {https://aclanthology.org/2020.aacl-main.75} {More data, more
  relations, more context and more openness: A review and outlook for relation
  extraction}.
\newblock In \emph{Proceedings of the 1st Conference of the Asia-Pacific
  Chapter of the Association for Computational Linguistics and the 10th
  International Joint Conference on Natural Language Processing}, pages
  745--758, Suzhou, China. Association for Computational Linguistics.

\bibitem[{Han et~al.(2018)Han, Zhu, Yu, Wang, Yao, Liu, and
  Sun}]{han-etal-2018-fewrel}
Xu~Han, Hao Zhu, Pengfei Yu, Ziyun Wang, Yuan Yao, Zhiyuan Liu, and Maosong
  Sun. 2018.
\newblock \href {https://doi.org/10.18653/v1/D18-1514} {{F}ew{R}el: A
  large-scale supervised few-shot relation classification dataset with
  state-of-the-art evaluation}.
\newblock In \emph{Proceedings of the 2018 Conference on Empirical Methods in
  Natural Language Processing}, pages 4803--4809, Brussels, Belgium.
  Association for Computational Linguistics.

\bibitem[{Hendrickx et~al.(2010)Hendrickx, Kim, Kozareva, Nakov,
  {\'O}~S{\'e}aghdha, Pad{\'o}, Pennacchiotti, Romano, and
  Szpakowicz}]{hendrickx-etal-2010-semeval}
Iris Hendrickx, Su~Nam Kim, Zornitsa Kozareva, Preslav Nakov, Diarmuid
  {\'O}~S{\'e}aghdha, Sebastian Pad{\'o}, Marco Pennacchiotti, Lorenza Romano,
  and Stan Szpakowicz. 2010.
\newblock \href {https://aclanthology.org/S10-1006} {{S}em{E}val-2010 task 8:
  Multi-way classification of semantic relations between pairs of nominals}.
\newblock In \emph{Proceedings of the 5th International Workshop on Semantic
  Evaluation}, pages 33--38, Uppsala, Sweden. Association for Computational
  Linguistics.

\bibitem[{Hendrycks et~al.(2017)Hendrycks, Gimpel, and
  Gimpel}]{DBLP:conf/iclr/HendrycksG17}
Dan Hendrycks, Kevin Gimpel, and Kevin Gimpel. 2017.
\newblock \href {https://openreview.net/forum?id=Hkg4TI9xl} {A baseline for
  detecting misclassified and out-of-distribution examples in neural networks}.
\newblock In \emph{5th International Conference on Learning Representations,
  {ICLR} 2017, Toulon, France, April 24-26, 2017, Conference Track
  Proceedings}. OpenReview.net.

\bibitem[{Hu et~al.(2020)Hu, Wen, Xu, Zhang, and Yu}]{hu-etal-2020-selfore}
Xuming Hu, Lijie Wen, Yusong Xu, Chenwei Zhang, and Philip Yu. 2020.
\newblock \href {https://doi.org/10.18653/v1/2020.emnlp-main.299} {{S}elf{ORE}:
  Self-supervised relational feature learning for open relation extraction}.
\newblock In \emph{Proceedings of the 2020 Conference on Empirical Methods in
  Natural Language Processing (EMNLP)}, pages 3673--3682, Online. Association
  for Computational Linguistics.

\bibitem[{Larson et~al.(2019)Larson, Mahendran, Peper, Clarke, Lee, Hill,
  Kummerfeld, Leach, Laurenzano, Tang, and Mars}]{larson-etal-2019-evaluation}
Stefan Larson, Anish Mahendran, Joseph~J. Peper, Christopher Clarke, Andrew
  Lee, Parker Hill, Jonathan~K. Kummerfeld, Kevin Leach, Michael~A. Laurenzano,
  Lingjia Tang, and Jason Mars. 2019.
\newblock \href {https://doi.org/10.18653/v1/D19-1131} {An evaluation dataset
  for intent classification and out-of-scope prediction}.
\newblock In \emph{Proceedings of the 2019 Conference on Empirical Methods in
  Natural Language Processing and the 9th International Joint Conference on
  Natural Language Processing (EMNLP-IJCNLP)}, pages 1311--1316, Hong Kong,
  China. Association for Computational Linguistics.

\bibitem[{Liang et~al.(2018)Liang, Li, and Srikant}]{liang2018enhancing}
Shiyu Liang, Yixuan Li, and R.~Srikant. 2018.
\newblock \href {https://openreview.net/forum?id=H1VGkIxRZ} {Enhancing the
  reliability of out-of-distribution image detection in neural networks}.
\newblock In \emph{International Conference on Learning Representations}.

\bibitem[{Lin and Xu(2019)}]{lin-xu-2019-deep}
Ting-En Lin and Hua Xu. 2019.
\newblock \href {https://doi.org/10.18653/v1/P19-1548} {Deep unknown intent
  detection with margin loss}.
\newblock In \emph{Proceedings of the 57th Annual Meeting of the Association
  for Computational Linguistics}, pages 5491--5496, Florence, Italy.
  Association for Computational Linguistics.

\bibitem[{Lin et~al.(2015)Lin, Liu, Sun, Liu, and
  Zhu}]{10.5555/2886521.2886624}
Yankai Lin, Zhiyuan Liu, Maosong Sun, Yang Liu, and Xuan Zhu. 2015.
\newblock Learning entity and relation embeddings for knowledge graph
  completion.
\newblock In \emph{Proceedings of the Twenty-Ninth AAAI Conference on
  Artificial Intelligence}, AAAI'15, page 2181–2187. AAAI Press.

\bibitem[{Liu et~al.(2020)Liu, Wang, Owens, and
  Li}]{WeitangLiu2020EnergybasedOD}
Weitang Liu, Xiaoyun Wang, John~D. Owens, and Yixuan Li. 2020.
\newblock Energy-based out-of-distribution detection.
\newblock \emph{neural information processing systems}.

\bibitem[{Ma et~al.(2021)Ma, Gui, Li, Zhang, Huang, and
  Zhou}]{ma-etal-2021-sent}
Ruotian Ma, Tao Gui, Linyang Li, Qi~Zhang, Xuanjing Huang, and Yaqian Zhou.
  2021.
\newblock \href {https://doi.org/10.18653/v1/2021.acl-long.484} {{SENT}:
  {S}entence-level distant relation extraction via negative training}.
\newblock In \emph{Proceedings of the 59th Annual Meeting of the Association
  for Computational Linguistics and the 11th International Joint Conference on
  Natural Language Processing (Volume 1: Long Papers)}, pages 6201--6213,
  Online. Association for Computational Linguistics.

\bibitem[{Madotto et~al.(2018)Madotto, Wu, and
  Fung}]{madotto-etal-2018-mem2seq}
Andrea Madotto, Chien-Sheng Wu, and Pascale Fung. 2018.
\newblock \href {https://doi.org/10.18653/v1/P18-1136} {{M}em2{S}eq:
  Effectively incorporating knowledge bases into end-to-end task-oriented
  dialog systems}.
\newblock In \emph{Proceedings of the 56th Annual Meeting of the Association
  for Computational Linguistics (Volume 1: Long Papers)}, pages 1468--1478,
  Melbourne, Australia. Association for Computational Linguistics.

\bibitem[{Nguyen et~al.(2015)Nguyen, Yosinski, and Clune}]{nguyen2015deep}
Anh Nguyen, Jason Yosinski, and Jeff Clune. 2015.
\newblock Deep neural networks are easily fooled: High confidence predictions
  for unrecognizable images.
\newblock In \emph{Proceedings of the IEEE conference on computer vision and
  pattern recognition}, pages 427--436.

\bibitem[{Recht et~al.(2019)Recht, Roelofs, Schmidt, and
  Shankar}]{recht2019imagenet}
Benjamin Recht, Rebecca Roelofs, Ludwig Schmidt, and Vaishaal Shankar. 2019.
\newblock Do imagenet classifiers generalize to imagenet?
\newblock In \emph{International Conference on Machine Learning}, pages
  5389--5400. PMLR.

\bibitem[{Ryu et~al.(2018)Ryu, Koo, Yu, and Lee}]{ryu-etal-2018-domain}
Seonghan Ryu, Sangjun Koo, Hwanjo Yu, and Gary~Geunbae Lee. 2018.
\newblock \href {https://doi.org/10.18653/v1/D18-1077} {Out-of-domain detection
  based on generative adversarial network}.
\newblock In \emph{Proceedings of the 2018 Conference on Empirical Methods in
  Natural Language Processing}, pages 714--718, Brussels, Belgium. Association
  for Computational Linguistics.

\bibitem[{Salton and Buckley(1987)}]{10.5555/866292}
Gerard Salton and Chris Buckley. 1987.
\newblock Term weighting approaches in automatic text retrieval.
\newblock Technical report, USA.

\bibitem[{Scheirer et~al.(2013)Scheirer, de~Rezende~Rocha, Sapkota, and
  Boult}]{6365193}
Walter~J. Scheirer, Anderson de~Rezende~Rocha, Archana Sapkota, and Terrance~E.
  Boult. 2013.
\newblock \href {https://doi.org/10.1109/TPAMI.2012.256} {Toward open set
  recognition}.
\newblock \emph{IEEE Transactions on Pattern Analysis and Machine
  Intelligence}, 35(7):1757--1772.

\bibitem[{Shu et~al.(2017)Shu, Xu, and Liu}]{shu-etal-2017-doc}
Lei Shu, Hu~Xu, and Bing Liu. 2017.
\newblock \href {https://doi.org/10.18653/v1/D17-1314} {{DOC}: Deep open
  classification of text documents}.
\newblock In \emph{Proceedings of the 2017 Conference on Empirical Methods in
  Natural Language Processing}, pages 2911--2916, Copenhagen, Denmark.
  Association for Computational Linguistics.

\bibitem[{Thulasidasan et~al.(2019{\natexlab{a}})Thulasidasan, Chennupati,
  Bilmes, Bhattacharya, and Michalak}]{NEURIPS2019_36ad8b5f}
Sunil Thulasidasan, Gopinath Chennupati, Jeff~A Bilmes, Tanmoy Bhattacharya,
  and Sarah Michalak. 2019{\natexlab{a}}.
\newblock \href
  {https://proceedings.neurips.cc/paper/2019/file/36ad8b5f42db492827016448975cc22d-Paper.pdf}
  {On mixup training: Improved calibration and predictive uncertainty for deep
  neural networks}.
\newblock In \emph{Advances in Neural Information Processing Systems},
  volume~32. Curran Associates, Inc.

\bibitem[{Thulasidasan et~al.(2019{\natexlab{b}})Thulasidasan, Chennupati,
  Bilmes, Bhattacharya, and Michalak}]{thulasidasan2019mixup}
Sunil Thulasidasan, Gopinath Chennupati, Jeff~A Bilmes, Tanmoy Bhattacharya,
  and Sarah Michalak. 2019{\natexlab{b}}.
\newblock On mixup training: Improved calibration and predictive uncertainty
  for deep neural networks.
\newblock \emph{Advances in Neural Information Processing Systems}, 32.

\bibitem[{Wang et~al.(2016)Wang, Cao, de~Melo, and
  Liu}]{wang-etal-2016-relation}
Linlin Wang, Zhu Cao, Gerard de~Melo, and Zhiyuan Liu. 2016.
\newblock \href {https://doi.org/10.18653/v1/P16-1123} {Relation classification
  via multi-level attention {CNN}s}.
\newblock In \emph{Proceedings of the 54th Annual Meeting of the Association
  for Computational Linguistics (Volume 1: Long Papers)}, pages 1298--1307,
  Berlin, Germany. Association for Computational Linguistics.

\bibitem[{Wu and He(2019)}]{DBLP:journals/corr/abs-1905-08284}
Shanchan Wu and Yifan He. 2019.
\newblock \href {http://arxiv.org/abs/1905.08284} {Enriching pre-trained
  language model with entity information for relation classification}.
\newblock \emph{CoRR}, abs/1905.08284.

\bibitem[{Xiong et~al.(2017)Xiong, Power, and Callan}]{xiong2017explicit}
Chenyan Xiong, Russell Power, and Jamie Callan. 2017.
\newblock Explicit semantic ranking for academic search via knowledge graph
  embedding.
\newblock In \emph{Proceedings of the 26th international conference on world
  wide web}, pages 1271--1279.

\bibitem[{Xu et~al.(2020)Xu, He, Yan, Liu, Liu, and Xu}]{xu-etal-2020-deep}
Hong Xu, Keqing He, Yuanmeng Yan, Sihong Liu, Zijun Liu, and Weiran Xu. 2020.
\newblock \href {https://doi.org/10.18653/v1/2020.coling-main.125} {A deep
  generative distance-based classifier for out-of-domain detection with
  mahalanobis space}.
\newblock In \emph{Proceedings of the 28th International Conference on
  Computational Linguistics}, pages 1452--1460, Barcelona, Spain (Online).
  International Committee on Computational Linguistics.

\bibitem[{Yan et~al.(2020)Yan, Fan, Li, Liu, Zhang, Wu, and
  Lam}]{yan-etal-2020-unknown}
Guangfeng Yan, Lu~Fan, Qimai Li, Han Liu, Xiaotong Zhang, Xiao-Ming Wu, and
  Albert~Y.S. Lam. 2020.
\newblock \href {https://doi.org/10.18653/v1/2020.acl-main.99} {Unknown intent
  detection using {G}aussian mixture model with an application to zero-shot
  intent classification}.
\newblock In \emph{Proceedings of the 58th Annual Meeting of the Association
  for Computational Linguistics}, pages 1050--1060, Online. Association for
  Computational Linguistics.

\bibitem[{Yu et~al.(2019)Yu, Qin, Liu, Zhao, Wang, and Chen}]{ijcai2019p0583}
Fuxun Yu, Zhuwei Qin, Chenchen Liu, Liang Zhao, Yanzhi Wang, and Xiang Chen.
  2019.
\newblock \href {https://doi.org/10.24963/ijcai.2019/583} {Interpreting and
  evaluating neural network robustness}.
\newblock In \emph{Proceedings of the Twenty-Eighth International Joint
  Conference on Artificial Intelligence, {IJCAI-19}}, pages 4199--4205.
  International Joint Conferences on Artificial Intelligence Organization.

\bibitem[{Zeng et~al.(2021)Zeng, He, Yan, Liu, Wu, Xu, Jiang, and
  Xu}]{zeng-etal-2021-modeling}
Zhiyuan Zeng, Keqing He, Yuanmeng Yan, Zijun Liu, Yanan Wu, Hong Xu, Huixing
  Jiang, and Weiran Xu. 2021.
\newblock \href {https://doi.org/10.18653/v1/2021.acl-short.110} {Modeling
  discriminative representations for out-of-domain detection with supervised
  contrastive learning}.
\newblock In \emph{Proceedings of the 59th Annual Meeting of the Association
  for Computational Linguistics and the 11th International Joint Conference on
  Natural Language Processing (Volume 2: Short Papers)}, pages 870--878,
  Online. Association for Computational Linguistics.

\bibitem[{Zhan and Zhao(2020)}]{Zhan_Zhao_2020}
Junlang Zhan and Hai Zhao. 2020.
\newblock \href {https://doi.org/10.1609/aaai.v34i05.6497} {Span model for open
  information extraction on accurate corpus}.
\newblock \emph{Proceedings of the AAAI Conference on Artificial Intelligence},
  34(05):9523--9530.

\bibitem[{Zhan et~al.(2021)Zhan, Liang, Liu, Fan, Wu, and
  Lam}]{zhan-etal-2021-scope}
Li-Ming Zhan, Haowen Liang, Bo~Liu, Lu~Fan, Xiao-Ming Wu, and Albert~Y.S. Lam.
  2021.
\newblock \href {https://doi.org/10.18653/v1/2021.acl-long.273} {Out-of-scope
  intent detection with self-supervision and discriminative training}.
\newblock In \emph{Proceedings of the 59th Annual Meeting of the Association
  for Computational Linguistics and the 11th International Joint Conference on
  Natural Language Processing (Volume 1: Long Papers)}, pages 3521--3532,
  Online. Association for Computational Linguistics.

\bibitem[{Zhang et~al.(2021)Zhang, Xu, and Lin}]{zhang2021deep}
Hanlei Zhang, Hua Xu, and Ting-En Lin. 2021.
\newblock Deep open intent classification with adaptive decision boundary.
\newblock In \emph{Proceedings of the AAAI Conference on Artificial
  Intelligence}, volume~35, pages 14374--14382.

\bibitem[{Zhang et~al.(2018)Zhang, Qi, and
  Manning}]{DBLP:journals/corr/abs-1809-10185}
Yuhao Zhang, Peng Qi, and Christopher~D. Manning. 2018.
\newblock \href {http://arxiv.org/abs/1809.10185} {Graph convolution over
  pruned dependency trees improves relation extraction}.
\newblock \emph{CoRR}, abs/1809.10185.

\bibitem[{Zhang et~al.(2017)Zhang, Zhong, Chen, Angeli, and
  Manning}]{zhang-etal-2017-position}
Yuhao Zhang, Victor Zhong, Danqi Chen, Gabor Angeli, and Christopher~D.
  Manning. 2017.
\newblock \href {https://doi.org/10.18653/v1/D17-1004} {Position-aware
  attention and supervised data improve slot filling}.
\newblock In \emph{Proceedings of the 2017 Conference on Empirical Methods in
  Natural Language Processing}, pages 35--45, Copenhagen, Denmark. Association
  for Computational Linguistics.

\bibitem[{Zhao et~al.(2021)Zhao, Gui, Zhang, and
  Zhou}]{zhao-etal-2021-relation}
Jun Zhao, Tao Gui, Qi~Zhang, and Yaqian Zhou. 2021.
\newblock \href {https://doi.org/10.18653/v1/2021.emnlp-main.765} {A
  relation-oriented clustering method for open relation extraction}.
\newblock In \emph{Proceedings of the 2021 Conference on Empirical Methods in
  Natural Language Processing}, pages 9707--9718, Online and Punta Cana,
  Dominican Republic. Association for Computational Linguistics.

\bibitem[{Zheng et~al.(2020)Zheng, Chen, and
  Huang}]{10.1109/TASLP.2020.2983593}
Yinhe Zheng, Guanyi Chen, and Minlie Huang. 2020.
\newblock \href {https://doi.org/10.1109/TASLP.2020.2983593} {Out-of-domain
  detection for natural language understanding in dialog systems}.
\newblock \emph{IEEE/ACM Trans. Audio, Speech and Lang. Proc.}, 28:1198–1209.

\bibitem[{Zhu et~al.(2019)Zhu, Lin, Liu, Fu, Chua, and
  Sun}]{zhu-etal-2019-graph}
Hao Zhu, Yankai Lin, Zhiyuan Liu, Jie Fu, Tat-Seng Chua, and Maosong Sun. 2019.
\newblock \href {https://doi.org/10.18653/v1/P19-1128} {Graph neural networks
  with generated parameters for relation extraction}.
\newblock In \emph{Proceedings of the 57th Annual Meeting of the Association
  for Computational Linguistics}, pages 1331--1339, Florence, Italy.
  Association for Computational Linguistics.

\end{thebibliography}
\bibliographystyle{acl_natbib}

\appendix

\section{Appendix}
\label{sec:appendix}
\subsection{Tf-idf statistic}
\label{sec:app-tf}
We consider a token $w_i$ to contribute significantly to a known relation $y\in \mathcal{K}$ if it occurs frequently in the instances of relation $y$ and rarely in the instances of other relations. \texttt{Tf-idf} statistic \cite{10.5555/866292} can well characterize this property.  Specifically, \texttt{Tf-idf} consists of term frequency and inverse document frequency. The term frequency $tf(w_i,y)$ describes how often a token $w_i$ appears in the instances of relation $y$: 
\begin{gather}
    tf(w_i,y)=\frac{n(w_i,y)}{\sum_{w_j\in\mathcal{V}}n(w_j,y)},
\end{gather}
where $n(w_i,y)$ denotes the number of times the token $w_i$ appears in the instances of relation $y$. Obviously, some tokens (e.g., the stop words) have high $tf$ values in different relational instances. However, they do not contribute to the relational semantics. The inverse document frequency describes whether the token $w_i$ appears only in the instances of specific relations:
\begin{gather}
    idf(w_i)=log\frac{|\mathcal{K}|}{|\{y:n(w_i,y)\neq0\}|},
\end{gather}
where $|\mathcal{K}|$ denotes total number of known relations and $|\{y:n(w_i,y)\neq0\}|$ denotes the number of known relations that token $w_i$ appears in their instances. Finally, we calculate $t(w_i,y)$ as follows:
\begin{gather}
    t(w_i,y)=tf(w_i,y)\times idf(w_i).
\end{gather}
The \texttt{tf-idf} statistic $t(w_i,y)$ measures the contribution of token $w_i$ to the relation semantics of $y$. We calculate and store the statistics based on the entire training set $\mathcal{D}_{\text{train}}$ before the training loop start. During the training, the statistic of each token in the vocabulary is fixed.

\subsection{How to Deal With Sub-tokens?}
\label{sec:app-sub}
BERT adopts BPE encoding to construct vocabularies. While most tokens are still single tokens, rare tokens are tokenized into sub-tokens. In this section, we introduce how to deal with sub-tokens when performing the substitution. First, the \texttt{tf-idf} statistics and the dependency scores are calculated at the token level and require no additional process. If a token consists of $n$ sub-tokens, we calculate its attribution score by summing the scores of all its sub-tokens. In addition, the misleading token of this token is only selected from the tokens that also have $n$ sub-tokens according to $\mathop{\arg\max}_{w_j\in \mathcal{V}_n}\sum_{k=1}^n\nabla_{\bm{w}_{i,k}}s_\theta(x)\cdot \bm{w}_{j,k}$. $\mathcal{V}_n$ denotes a vocabulary, in which all tokens consist of $n$ sub-tokens. $\bm{w}_{i,k}$ denotes the embedding of the $k^{\text{th}}$ sub-token of the token $w_i$.

\subsection{Compared Methods}
\label{sec:app-baseline}
To validate the effectiveness of the proposed method, we compare our method with mainstream open-set classification methods.

\noindent\textbf{MSP} \cite{DBLP:conf/iclr/HendrycksG17}. MSP assumes that correctly classified instances tend to have greater maximum softmax probability than samples of unknown classes. Therefore, the maximum softmax probability is used as the detection score.

\noindent\textbf{DOC} \cite{shu-etal-2017-doc}. DOC builds a 1-vs-rest layer containing $m$ binary sigmoid classifiers for $m$ known classes. The maximum probability of m binary classifiers is used as the detection score.

\noindent\textbf{ODIN} \cite{liang2018enhancing}. Based on MSP, ODIN uses temperature scaling and small perturbations to separate the softmax score distributions between samples of known and unknown classes.
	
\noindent\textbf{MixUp} \cite{thulasidasan2019mixup}. MixUp trains the model on convexly combined pairs of instances, which is effective to calibrate the softmax scores.

\noindent\textbf{Energy} \cite{WeitangLiu2020EnergybasedOD}. Instead of maximum softmax probability, this method uses the free energy $E(x) = - \log \sum_{k=1}^K e^{f_k(x)}$ as the detection score of the unknown data.
	
\noindent\textbf{Convex} \cite{zhan-etal-2021-scope}. The method learns a more discriminative representation by generating synthetic outliers using inlier features.

%The method constructs pseudo outliers during training, by generating synthetic outliers using inlier features in a self-supervision manner.
% The pseudo outliers are used to learn more discriminative representation.

\noindent\textbf{SCL} \cite{zeng-etal-2021-modeling}. SCL proposes a supervised contrastive learning objective, learning a more discriminative representation for unknown data detection.

\subsection{Relations comprising the datasets}
\label{sec:rel}
In this subsection, we present the known relations contained in the training set, the unknown relations included in the validation set, and the unknown relations present in the test set, as shown in Table \ref{tab:rel}.
    \begin{table}[h]
            \centering
            \resizebox{\linewidth}{!}{
            \begin{tabular}{l}
            \toprule
            \textbf{Relations in FewRel}:\\
            \textbf{Training Set}: P241, P22, P460, P4552, P140, P39, P118, P674, P361,\\ P1408, P410, P931, P1344, P1303, P1877, P407, P105, P3450, P991, P800, \\P40, P551, P750, P106, P364, P706, P127, P150, P131, P159, P264, P102,\\ P974, P84, P155, P31, P740, P26, P177, P206\\
            \textbf{Validation Set}: P135, P403, P1001, P59, P25, P412, P413, P136, P178,\\ P1346, P921, P123, P17, P1435, P306, P641, P101, P495, P466, P58\\
            \textbf{Testing Set}: P57, P6, P2094, P1923, P463, P1411, P710, P176, P355,\\ P400, P449, P276, P156, P137, P27, P527, P175, P3373, P937, P86\\
            \midrule
            \textbf{Relations in FewRel}:\\
            \textbf{Training Set}:     per:stateorprovince\_of\_death, org:shareholders,\\
    org:alternate\_names, per:country\_of\_birth, org:city\_of\_headquarters,\\
    per:age, per:cities\_of\_residence, per:children, org:members, org:founded\\
    per:title, org:website, per:alternate\_names, org:country\_of\_headquarters,\\
    per:stateorprovinces\_of\_residence, per:cause\_of\_death, per:charges\\
    org:political\_religious\_affiliation, org:parents, org:dissolved, per:spouse,\\
            \textbf{Validation Set}: org:subsidiaries, per:city\_of\_birth, 
            per:date\_of\_death,\\
            per:stateorprovince\_of\_birth, per:employee\_of, org:member\_of, per:origin,\\
    per:date\_of\_birth, per:countries\_of\_residence, org:founded\_by\\
            \textbf{Testing Set}:     org:stateorprovince\_of\_headquarters, per:country\_of\_death,\\
    per:religion, per:city\_of\_death, org:number\_of\_employees\_members,\\
    per:parents, per:schools\_attended, per:siblings, per:other\_family,\\
    org:top\_members\_employees,
    no\_relation\\
            \bottomrule
            \end{tabular}
            }
            \caption{Relations comprising each dataset.}
            \label{tab:rel}
        \end{table}   
    \begin{table*}[h]
            \centering
            \resizebox{\linewidth}{!}{
            \begin{tabular}{l}
            \toprule
            \textbf{Relation}: Instrument (musical instrument that a person plays) \\
            \textbf{Tokens with top 10 tf-idf statistics}: bass, saxophone, guitar, player, trumpet, trombone, composer, drums, organ, cello\\
            \textbf{Original Training Instance}: In 1961, McIntosh composed a \textcolor{red}{\textbf{song}} for [\textcolor{red}{\textbf{trumpet}}]$_{tail}$ legend [Howard Mcghee]$_{head}$.\\
            \textbf{Synthesized Negative Instance}: In 1961, McIntosh composed a \textcolor{red}{\textbf{verse}} for [\textcolor{red}{\textbf{Mississippi}}]$_{tail}$ legend [Howard Mcghee]$_{head}$.\\
            \midrule
            \textbf{Relation}: Spouse (a husband or wife, considered in relation to their partner.) \\
            \textbf{Tokens with top 10 tf-idf statistics}: wife, husband, married, survived, died, grandchildren, children, heidi, sons, robert\\
            \textbf{Original Training Instance}: ``[his]$_{head}$ \textcolor{red}{\textbf{family}} was at his bedside'', his \textcolor{red}{\textbf{wife}}, [Barbara Washburn]$_{tail}$, said Thursday.\\
            \textbf{Synthesized Negative Instance}: ``[his]$_{head}$ \textcolor{red}{\textbf{friend}} was at his bedside'', his \textcolor{red}{\textbf{captain}}, [Barbara Washburn]$_{tail}$, said Thursday.\\
            \bottomrule
            \end{tabular}
            }
            \caption{Case study of the proposed negative samples synthesis method. The relation semantics between the given entity pair is completely erased by substituting only 2 tokens (tokens in red).}
            \label{tab:case}
        \end{table*}   
\subsection{Additional Results}
\noindent\textbf{Classification Accuracy}: One of our key claims is that the proposed method achieves state-of-the-art SOTA detection without compromising the classification of known relations. In this section, we provide an additional \textbf{ACC} metric, in which only the instances of $n$ known relations are used to calculate the classification accuracy. The metric exactly indicates whether \texttt{NOTA} detection impairs the classification of known relations. From tab. \ref{tab:known} we can observe that our method is comparable to the existing method, which supports the key claim at the beginning of the paragraph.

        \begin{table}[h]
            \centering
            \resizebox{\columnwidth}{!}{
            \begin{tabular}{l cc}
            \toprule
            \textbf{Method}& FewRel & TACRED\\
            \midrule
            MSP \cite{DBLP:conf/iclr/HendrycksG17}& 93.13$_{0.41}$ & 94.77$_{0.98}$\\
            DOC \cite{shu-etal-2017-doc}& 93.25$_{0.17}$ & 93.70$_{0.16}$\\
            ODIN \cite{liang2018enhancing}& 93.11$_{0.38}$ & 94.88$_{0.57}$\\
            MixUp \cite{thulasidasan2019mixup}& 93.19$_{0.41}$ & 94.37$_{1.28}$\\
            Energy \cite{WeitangLiu2020EnergybasedOD}& 93.36$_{0.18}$ & 94.97$_{0.54}$\\
            Convex \cite{zhan-etal-2021-scope}& 91.97$_{0.96}$ & 93.10$_{0.21}$\\
            SCL \cite{zeng-etal-2021-modeling}& 93.45$_{0.08}$ & 95.20$_{0.50}$\\
            \textbf{Ours} & \textbf{93.50}$_{0.37}$ & \textbf{95.53}$_{0.17}$\\
            \bottomrule
            \end{tabular}
            }
            \caption{The results of \textbf{ACC} on $n$ known relations. The subscript represents the corresponding standard deviation (e.g., 93.50$_{0.37}$ indicates 93.50$\pm0.37$).}
            \label{tab:known}
        \end{table}   
        
\noindent\textbf{Two Real Substitution Cases}: To intuitively show the effectiveness of the proposed synthesis method, we conduct a case study based on the ``Instrument'' relation from FewRel and the ``Spouse'' relation from TACRED. 
The tokens with top-10 \texttt{tf-idf} statistics and a substitution case of each relation are shown in tab. \ref{tab:case}, from which we can observe that: (1) the tokens with high \texttt{tf-idf} statistics have a strong semantic association with the target relation (such as Instrument-bass, Spouse-wife). (2) By substituting only two critical tokens in original training instances, the target relation is completely erased.

 %From table \ref{tab:case} we can observe that: (1) The relation-significant words have a strong semantic association with the corresponding relations (such as Instrument-bass, Spouse-wife). This strong semantic association is the basis of the success of our method. (2) With MLM, we can synthesize fluent and grammatically correct negative samples. In the examples in table \ref{tab:case}, we completely erase the relation semantics between the given entity pairs by substituting only 2 words.

\end{document}